%% file: main.tex
\documentclass[10pt,twocolumn,letterpaper]{article}

\usepackage[pagenumbers]{wacv} 


\definecolor{wacvblue}{rgb}{0.21,0.49,0.74}
\usepackage[pagebackref,breaklinks,colorlinks,allcolors=wacvblue]{hyperref}

\def\wacvPaperID{1146} 
\def\confName{WACV}
\def\confYear{2026}

\makeatletter
\newcommand*\bigcdot{\mathpalette\bigcdot@{2}}
\newcommand*\bigcdot@[2]{\mathbin{\vcenter{\hbox{\scalebox{#2}{$\m@th#1\bullet$}}}}}
\makeatother

\title{JOCA: Task-Driven \underline{J}oint \underline{O}ptimisation of \underline{C}amera\\Hardware and \underline{A}daptive Camera Control Algorithms}

\author{Chengyang Yan \qquad Mitch Bryson \qquad Donald G. Dansereau\\
Australian Centre for Robotics, School of Aerospace, Mechanical and Mechatronic Engineering, \\ The University of Sydney\\
{\tt\small {chengyang.yan, mitch.bryson, donald.dansereau}@sydney.edu.au}
}

\begin{document}
\maketitle
\input{sec/0_abstract}
\input{sec/1_intro}
\input{sec/2_lit}
\input{sec/3_image_formation}
\input{sec/4_ACC}
\input{sec/5_Optimization}
\input{sec/6_Experiment}
\input{sec/7_Conclusion}
\newpage
{
    \small
    \bibliographystyle{ieeenat_fullname}
    \bibliography{main}
}

\clearpage \appendix \input{supplementary}

\end{document}

%% file: sec/0_abstract.tex
\begin{abstract}
The quality of captured images strongly influences the performance of downstream perception tasks. Recent works on co-designing camera systems with perception tasks have shown improved task performance. However, most prior approaches focus on optimising fixed camera parameters set at manufacturing, while many parameters, such as exposure settings, require adaptive control at runtime. This paper introduces a method that jointly optimises camera hardware and adaptive camera control algorithms with downstream vision tasks. We present a unified optimisation framework that integrates gradient-based and derivative-free methods, enabling support for both continuous and discrete parameters, non-differentiable image formation processes, and neural network-based adaptive control algorithms. To address non-differentiable effects such as motion blur, we propose DF-Grad, a hybrid optimisation strategy that trains adaptive control networks using signals from a derivative-free optimiser alongside unsupervised task-driven learning. Experiments show that our method outperforms baselines that optimise static and dynamic parameters separately, particularly under challenging conditions such as low light and fast motion. These results demonstrate that jointly optimising hardware parameters and adaptive control algorithms improves perception performance and provides a unified approach to task-driven camera system design. Code is available via our project page at \url{https://roboticimaging.org/Projects/JOCA/}.
\end{abstract}

%% file: sec/1_intro.tex
\section{Introduction}
\label{sec:intro}

\begin{figure}
      \centering
      \includegraphics[width=\linewidth]{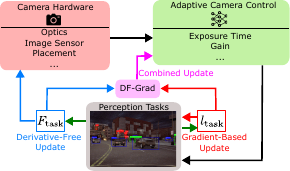}
      \caption{We introduce a novel end-to-end method that jointly optimises camera hardware parameters, adaptive camera control algorithms, and perception tasks to improve task performance. We combine a fitness function $F_{\text{task}}$ for a derivative-free optimiser and a loss function $l_{\text{task}}$ for a gradient-based optimiser using the proposed DF-Grad method to update neural network-based adaptive control algorithms, allowing them to learn from a non-differentiable image formation process. The method supports optimisation of static camera parameters that are continuous and discrete, as well as dynamic camera parameters, enabling task-aware and adaptive camera design.}
      \label{fig:key_figure}
\end{figure}

\hspace*{\parindent} Image quality directly determines the performance of downstream perception tasks. A well-designed camera hardware architecture preserves essential visual information, while adaptive imaging algorithms maintain quality under varying capture conditions. For example, an adaptive camera control (ACC) algorithm that dynamically adjusts exposure can reduce degradation caused by illumination changes or motion.

Recent works have proposed task-driven ACC algorithms, particularly for auto-exposure (AE), that optimise exposure based on task performance. Tomasi et al.~\cite{tomasi2021learned} trained an AE network for feature extraction, while Onzon et al.~\cite{onzon2021neural} jointly trained an AE network with an object detector. However, these methods assume fixed camera hardware and neglect the interaction between dynamic parameters and key image effects such as motion blur, which depends on exposure time and scene dynamics.

In parallel, several studies have explored joint optimisation of imaging systems and perception tasks, consistently outperforming traditional decoupled pipelines. Klinghoffer et al.~\cite{klinghoffer2023diser} introduced a reinforcement learning (RL)-based framework that jointly trains a camera-design network and a perception model, supporting both discrete and continuous parameters as well as non-differentiable simulations. Yan et al.~\cite{yan2025tacos} improved this process by directly optimising camera parameters with a derivative-free (DF) optimiser, improving efficiency while retaining support for non-differentiable components. Yet, these works do not account for the role of ACC algorithms in the joint optimisation process.

In this work, we present the first end-to-end framework that jointly optimises fixed camera hardware parameters, ACC algorithms, and downstream perception models (Fig.~\ref{fig:key_figure}). Since many camera parameters are inherently discrete, and many established large-scale simulators with realistic physics for dynamic scenes are non-differentiable, we design the method to support continuous and discrete parameters, as well as non-differentiable image formation processes.  Inspired by~\cite{yan2025tacos}, we employ a genetic algorithm (GA)~\cite{holland1992adaptation} to optimise hardware in a non-differentiable simulator, while using a gradient-based optimiser for perception tasks. Our key innovation is DF-Grad, a hybrid training strategy that combines gradient-based optimisation with supervision from a derivative-free optimiser to train the ACC algorithm. This enables ACC learning under realistic, non-differentiably rendered image effects.

We validate our method on camera design problems requiring dynamic adjustment of exposure parameters and study the influence of hardware and ACC algorithms on task performance. Experiments show that our joint optimisation scheme consistently outperforms prior methods, especially in challenging conditions with low illumination and strong motion blur. In summary, our contributions are:
\begin{itemize}[noitemsep]
\item We present the first end-to-end framework that jointly optimises camera hardware, adaptive camera control algorithms, and downstream perception tasks.
\item We introduce DF-Grad, a hybrid optimisation strategy integrating derivative-free and gradient-based methods to train deep learning-based ACC algorithms under non-differentiable rendering of image effects.
\item We demonstrate that our approach consistently outperforms prior methods in both standard and challenging conditions, with a detailed analysis of the impact of joint learning and DF-Grad.
\end{itemize}

We believe this work advances the automated design of intelligent imaging systems that emphasise task performance in dynamic environments.

%% file: sec/2_lit.tex
\section{Related Work}
\label{sec:related}
\begin{figure*}
        \centering
        \includegraphics[width=\linewidth]{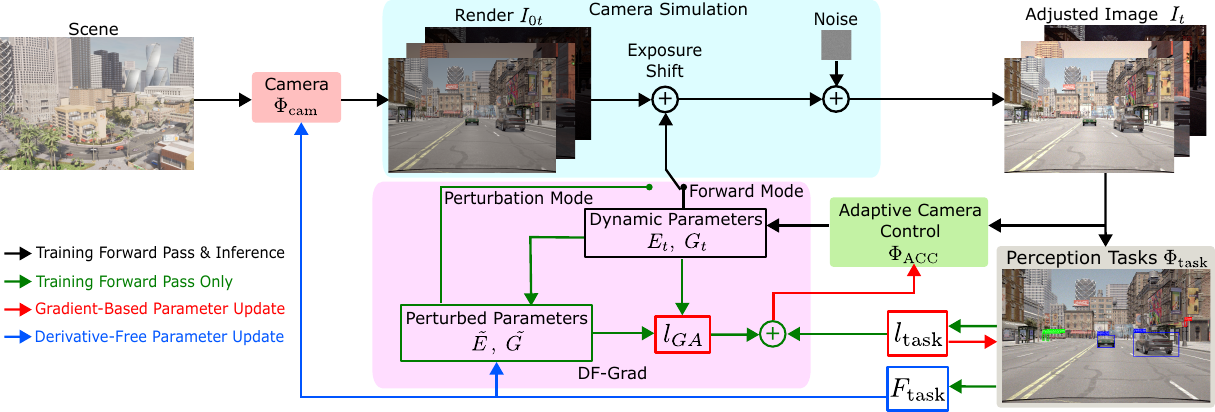}
        \caption{A virtual camera with configurable parameters is used to render images ($I_{0t}$) from scenes under dynamic illumination. In forward mode (standard mode for training and inference), the render is adjusted by dynamic parameters predicted by the ACC algorithm and augmented with physically realistic noise to produce the adjusted image ($I_t$), which is then evaluated by the perception task and used to predict dynamic parameters for the next frame. During training, camera parameters $\Phi_{\text{cam}}$ are optimised by GA using a task fitness function $F_{\text{task}}$ (blue arrow), while task model parameters $\Phi_{\text{task}}$ are updated by gradient descent on the task loss $l_{\text{task}}$ (red arrow). The ACC network parameters $\Phi_{\text{ACC}}$ are trained using the proposed DF-Grad method, which combines the task loss $l_{\text{task}}$ with a GA supervision loss $l_{\text{GA}}$. The latter measures the difference between ACC predictions and GA-perturbed parameters. The combined loss is used to update $\Phi_{\text{ACC}}$ via a gradient-based method. In perturbation mode (only for perturbation optimisation), the render is adjusted with GA-perturbed parameters and evaluated in the same way as in forward mode.}
\label{fig:method_overview}
\end{figure*}
\smallskip \noindent \textbf{Task-Driven Camera Hardware Design}\hspace{2pt} Designing cameras specifically for perception tasks has been shown to enhance performance. Early approaches renderers such as ISET~\cite{Wandell2024ISET3D, Wandell2024ISETCam} to prototype cameras and tune parameters based on task outcomes~\cite{blasinski2018optimizing, liu2019system, liu2019soft, weikl2021end, liu2023using}. These methods improved performance and reduced development costs, but relied on manual parameter tuning. More recent methods, including ours, enable automatic, end-to-end optimisation.

End-to-end camera design methods typically employ gradient-based optimisation with differentiable simulation, jointly optimising parameters and tasks. Applications include extending depth of field (DoF)~\cite{sitzmann2018end, sun2021end, yang2024curriculum}, object detection~\cite{tseng2019hyperparameter, tseng2021differentiable, robidoux2021end, cote2023differentiable}, image classification~\cite{chang2018hybrid, diamond2021dirty, zhang2022all, yang2023image}, depth estimation~\cite{he2018learning, chang2019deep, ikoma2021depth, baek2021polka}, HDR imaging~\cite{metzler2020deep, martel2020neural}, and motion deblurring~\cite{nguyen2022learning}. However, these methods typically rely on differentiable ray tracing over pre-captured datasets, limiting generalisability to tasks involving variable fields of view (FoV), resolutions, unconventional configurations, or multi-camera systems. Moreover, gradient-based optimisation cannot directly handle discrete parameters. In contrast, our approach leverages a derivative-free optimiser to support more flexible and complex camera design scenarios.

Alternative strategies have been explored to support non-differentiable simulators and discrete parameters. RL-based methods train neural networks to design cameras jointly with perception tasks~\cite{klinghoffer2023diser, hou2023optimizing}. More recently, Yan et al.~\cite{yan2025tacos} and Tiwary et al.~\cite{tiwary2025if} employed evolutionary algorithms to directly optimise camera and vision system parameters, bypassing large neural networks and enabling faster optimisation with less data. While effective and generalisable, these approaches do not address dynamic parameters.

Here, we extend joint design by simultaneously optimising hardware parameters and adaptive control algorithms that manage dynamic parameters at runtime.

\smallskip \noindent \textbf{Task-Driven Adaptive Camera Control}\hspace{2pt} ACC algorithms are widely used to adapt cameras to dynamic conditions. Many works control dynamic parameters, especially those related to exposure and image signal processing, based on task performance, targeting applications such as feature extraction for visual odometry~\cite{shim2014auto, tomasi2021learned, han2023camera}, object detection~\cite{stojkovic2023deep}, localisation~\cite{begin2022auto}, and homography estimation~\cite{lin2023reinforcement}. Other approaches jointly train neural ACC algorithms with perception tasks in an end-to-end manner, yielding improved performance~\cite{onzon2021neural, yoshimura2023dynamicisp, shopovska2023high}. Our method further co-designs camera hardware, enabling tighter integration between hardware, ACC algorithms, and downstream tasks.

Moreover, existing ACC methods neglect effects induced by dynamic parameter changes. Using the proposed DF-Grad, the network learns connections between parameters and effects, even when rendered non-differentiably.

\smallskip \noindent \textbf{Hybrid Gradient-Based and Derivative-Free Optimisation}\hspace{2pt} Prior work has combined derivative-free and gradient-based optimisation in other contexts. ERL~\cite{khadka2018evolution} and CERL~\cite{khadka2019collaborative} use evolutionary algorithms to generate policies for supervising gradient-trained agents. AutoAugment~\cite{cubuk2019autoaugment} applies evolutionary search to identify augmentation strategies that guide policy networks. To our knowledge, our method is the first to combine derivative-free and gradient-based optimisation for training ACC networks, enabling effective learning with non-differentiable image formation.

%% file: sec/3_image_formation.tex
\section{Image Formation}\label{sec:image_formation}
\hspace*{\parindent} We propose a joint optimisation method for camera hardware, ACC algorithms, and perception tasks, illustrated in Fig.~\ref{fig:method_overview}. This section outlines the image formation model used. In digital cameras, pixel intensity corresponds to the number of photons arriving at a pixel area ($\phi$) during exposure time ($E$), scaled by the camera gain ($G$):
\begin{equation}\label{eqn:image_formation}
I_{i, j} = E G \phi,
\end{equation}
where $I_{i,j}$ is the intensity at pixel location $(i, j)$ in image $I$.

We use a virtual camera to capture a simulated environment or pre-captured real-world images in this work. The images are calibrated against a physical camera with known dynamic parameters ($E_0$ and $G_0$), which are also employed during optimisation for intensity adjustment. The calibrated images serve as $I$ in~(\ref{eqn:image_formation}). We assume global shutters and add motion blur to the images based on $E$ and scene motion.

In operation, intensities at time $t$ are updated using the exposure ($E_t$) and gain ($G_t$) predicted by the ACC network:
\begin{equation}\label{eqn:image_scale}
I = I_0 \frac{E_t}{E_0} \frac{G_t}{G_0},
\end{equation}
where $I_0$ is the calibrated image. This process is the forward mode in Fig.~\ref{fig:method_overview}.

Image noise is a key phenomenon in image formation. As most camera simulators lack realistic noise models, we augment $I$ with synthetic noise based on the affine noise model~\cite{foi2009clipped}. This includes signal-dependent Poisson noise and thermal noise, calibrated with a physical camera following established procedures~\cite{wang2021multiplexed, yan2025tacos}. The same calibration parameters ($E_0$, $G_0$) are reused for calibrating the noise model. We then generalise this model to handle variable $E_t$ and $G_t$. The generalised pixel-level intensity variance is
\begin{equation}\label{eqn:noise_generlization}
\sigma^2 = \frac{G_t}{G_0} \sigma_p^2 I + \frac{G_t^2}{G_0^2} \sigma_r^2,
\end{equation}
where $\sigma_p$ and $\sigma_r$ denote the calibrated photon and thermal noise parameters, respectively, and $I$ is the image intensity from~(\ref{eqn:image_scale})~\cite{yan2025tacos}. Dark-current noise is omitted because the exposure times are short ($E_t \ll 1$ s) and operation is assumed to occur at room temperature~\cite{abarca2023cmos}. This noise model was proposed and validated in~\cite{yan2025tacos}.

Finally, pixel intensities are clipped to the sensor’s white level ($M_{\text{white}}$), the maximum measurable value. We use 8-bit images, where $M_{\text{white}} = 2^8 - 1$.

%% file: sec/4_ACC.tex

\section{Adaptive Camera Control Algorithm}\label{sec:AEC}
\hspace*{\parindent} We adopt and extend the NeuralAE model proposed in \cite{onzon2021neural} as the ACC algorithm to dynamically control two camera parameters: exposure time and gain. The architecture of the AE model is shown in the supplementary material.

We keep the global image feature branch and the semantic feature branch in the original model~\cite{onzon2021neural}. While the original method uses two cameras, we simplify the setup to a single camera. Image captured ($I_t$) is used to predict the exposure for the next frame ($I_{t+1}$) at every step. To improve temporal awareness, we modify the semantic feature branch by adding a parameter-awareness component. This component concatenates the dynamic parameters predicted for the previous frame with the image features extracted in the semantic feature branch, allowing the network to condition its predictions on prior settings. The outputs from both branches are summed and passed through a multi-layer perceptron (MLP), which we modify to predict both exposure time ($E_{\text{pred}}$) and gain ($G_{\text{pred}}$), rather than only exposure as in the original method. Architectural details follow~\cite{onzon2021neural}.

We then adjust $E_{\text{pred}}$ and $G_{\text{pred}}$ to account for varying pixel sizes. If the calibration camera has a pixel size $p_0$ and the optimiser proposes a new pixel size $p$, the corresponding calibrated image intensity ($I_0$) is scaled by the pixel area ratio $\frac{p^2}{p_0^2}$, assuming square pixels. To compensate for intensity differences induced by pixel sizes while preserving a high signal-to-noise ratio (SNR), we rescale the predictions with priority of minimising gain, since lower gain improves SNR as in~(\ref{eqn:noise_generlization}). The final exposure settings at timestep $t$ are:
\begin{equation}\label{eqn:exposure_scale} 
\begin{aligned} 
G_t &= \begin{cases} \max\left(\frac{p_0^2}{p^2} G_{\text{pred}}, 1\right) & \text{if } \frac{p^2}{p_0^2} > 1,\\ 
G_{\text{pred}} & \text{otherwise}, 
\end{cases} \\ 
E_t &= \begin{cases} \frac{p_0^2}{p^2} \frac{1}{G_t} E_{\text{pred}} G_{\text{pred}} & \text{if } \frac{p^2}{p_0^2} > 1,\\ 
\frac{p_0^2}{p^2} E_{\text{pred}} & \text{otherwise}. 
\end{cases} 
\end{aligned} 
\end{equation}

The resulting $E_t$ and $G_t$ are used as the final exposure settings in~(\ref{eqn:image_scale}), driving both rendering adjustments and image effects, including motion blur and image noise.

%% file: sec/5_Optimization.tex
\begin{table*}[!ht]
\centering
\caption{
Experiment results using synthetic images. We compare the parameters and performance of our method against existing methods that do not jointly optimise the camera, ACC algorithm, and perception task. Optimised camera parameters are indicated by \textcolor{green}{$\bigcdot$}, while human-designed camera parameters are denoted by $\bigcdot$. Our approach achieves slightly improved performance compared to methods that jointly optimise only the ACC algorithm and perception task under the scenario with calibrated noise level and motion blur. Under more challenging conditions with stronger motion blur and lower SNRs, our method significantly outperforms all existing approaches.
}
\resizebox{2\columnwidth}{!}{%
\begin{tabular}{ccccccccc}
\hline
\multirow{3}{*}{Scenario} & \multirow{3}{*}{Method} & \multicolumn{5}{c}{Camera Parameters} & \multicolumn{2}{c}{Performance} \\
\cline{3-9}
&  & Forward Position & Height & Focal Length & Sensor Size & Pixel Size &  Object Detect. & True Positive Ratio\\
& & x ($m$) & z ($m$) & $f$ (mm) & $w\times h$  (mm) & $p$ ($\mu$m) & mAP $\uparrow$ & $\text{TP}/\text{All@}180^{\circ}\uparrow$\\
\hline
\multirow{4}{*}{Calibrated} & FLIR + nuScenes~\cite{caesar2020nuscenes} + NeuralAE~\cite{onzon2021neural} (Gradient-Based) & 0.43 $\bigcdot$ & 1.65 $\bigcdot$ & 3.6 $\bigcdot$ & 6.2$\times$4.65 $\bigcdot$ & 1.55 $\bigcdot$ & \underline{0.318} & \underline{0.389}\\
& Basler + nuScenes~\cite{caesar2020nuscenes} + NeuralAE~\cite{onzon2021neural} (Gradient-Based) & 0.43 $\bigcdot$ & 1.65 $\bigcdot$ & 3.6 $\bigcdot$ & 4.8$\times$3.6 $\bigcdot$ & 3.75 $\bigcdot$ & 0.289 & 0.360 \\
& TaCOS~\cite{yan2025tacos} + AverageAE~\cite{ARM} & 2.43 \textcolor{green}{$\bigcdot$} & 1.7 \textcolor{green}{$\bigcdot$} & 10 \textcolor{green}{$\bigcdot$} & 14.44$\times$9.9 \textcolor{green}{$\bigcdot$} & 4.5 \textcolor{green}{$\bigcdot$} & 0.303 & 0.358\\
& Ours - TaCOS~\cite{yan2025tacos} + NeuralAE~\cite{onzon2021neural} (DF-Grad) & 1.35 \textcolor{green}{$\bigcdot$} & 1.3 \textcolor{green}{$\bigcdot$} & 5.19 \textcolor{green}{$\bigcdot$} & 7.37$\times$4.92 \textcolor{green}{$\bigcdot$} & 2.4 \textcolor{green}{$\bigcdot$} & \textbf{0.325} & \textbf{0.390}\\
\hline
\multirow{4}{*}{Noise$\times$10} & FLIR + nuScenes~\cite{caesar2020nuscenes} + NeuralAE~\cite{onzon2021neural} (Gradient-Based) & 0.43 $\bigcdot$ & 1.65 $\bigcdot$ & 3.6 $\bigcdot$ & 6.2$\times$4.65 $\bigcdot$ & 1.55 $\bigcdot$ & 0.263 & \textbf{0.382}\\
& Basler + nuScenes~\cite{caesar2020nuscenes} + NeuralAE~\cite{onzon2021neural} (Gradient-Based) & 0.43 $\bigcdot$ & 1.65 $\bigcdot$ & 3.6 $\bigcdot$ & 4.8$\times$3.6 $\bigcdot$ & 3.75 $\bigcdot$ & 0.286 & 0.334 \\
& TaCOS~\cite{yan2025tacos} + AverageAE~\cite{ARM} & 1.57 \textcolor{green}{$\bigcdot$} & 1.66 \textcolor{green}{$\bigcdot$} & 7.91 \textcolor{green}{$\bigcdot$} & 11.25$\times$7.03 \textcolor{green}{$\bigcdot$} & 5.86 \textcolor{green}{$\bigcdot$} & \underline{0.288} & 0.363\\
& Ours - TaCOS~\cite{yan2025tacos} + NeuralAE~\cite{onzon2021neural} (DF-Grad) & 2.07 \textcolor{green}{$\bigcdot$} & 1.52 \textcolor{green}{$\bigcdot$} & 7.64 \textcolor{green}{$\bigcdot$} & 11.34 $\times$7.13 \textcolor{green}{$\bigcdot$} & 5.86 \textcolor{green}{$\bigcdot$} & \textbf{0.296} & \underline{0.377}\\
\hline
\multirow{4}{*}{Noise$\times$20} & FLIR + nuScenes~\cite{caesar2020nuscenes} + NeuralAE~\cite{onzon2021neural} (Gradient-Based) & 0.43 $\bigcdot$ & 1.65 $\bigcdot$ & 3.6 $\bigcdot$ & 6.2$\times$4.65 $\bigcdot$ & 1.55 $\bigcdot$ & 0.234 & 0.330\\
& Basler + nuScenes~\cite{caesar2020nuscenes} + NeuralAE~\cite{onzon2021neural} (Gradient-Based) & 0.43 $\bigcdot$ & 1.65 $\bigcdot$ & 3.6 $\bigcdot$ & 4.8$\times$3.6 $\bigcdot$ & 3.75 $\bigcdot$ & \underline{0.276} & 0.358 \\
& TaCOS~\cite{yan2025tacos} + AverageAE~\cite{ARM} & 2.07 \textcolor{green}{$\bigcdot$} & 1.51 \textcolor{green}{$\bigcdot$} & 10 \textcolor{green}{$\bigcdot$} & 16.13$\times$12.04 \textcolor{green}{$\bigcdot$} & 7 \textcolor{green}{$\bigcdot$} & 0.245 & \underline{0.371}\\
& Ours - TaCOS~\cite{yan2025tacos} + NeuralAE~\cite{onzon2021neural} (DF-Grad) & 1.54 \textcolor{green}{$\bigcdot$} & 1.56 \textcolor{green}{$\bigcdot$} & 8.51 \textcolor{green}{$\bigcdot$} & 14.13$\times$7.45 \textcolor{green}{$\bigcdot$} & 3.45 \textcolor{green}{$\bigcdot$} & \textbf{0.289} & \textbf{0.381}\\
\hline
\multirow{4}{*}{Motion Blur$\times$2} & FLIR + nuScenes~\cite{caesar2020nuscenes} + NeuralAE~\cite{onzon2021neural} (Gradient-Based) & 0.43 $\bigcdot$ & 1.65 $\bigcdot$ & 3.6 $\bigcdot$ & 6.2$\times$4.65 $\bigcdot$ & 1.55 $\bigcdot$ & \underline{0.297} & 0.345\\
& Basler + nuScenes~\cite{caesar2020nuscenes} + NeuralAE~\cite{onzon2021neural} (Gradient-Based) & 0.43 $\bigcdot$ & 1.65 $\bigcdot$ & 3.6 $\bigcdot$ & 4.8$\times$3.6 $\bigcdot$ & 3.75 $\bigcdot$ & 0.259 & 0.358 \\
& TaCOS~\cite{yan2025tacos} + AverageAE~\cite{ARM} & 2.43 \textcolor{green}{$\bigcdot$} & 1.7 \textcolor{green}{$\bigcdot$} & 7.15 \textcolor{green}{$\bigcdot$} & 11.26$\times$5.98 \textcolor{green}{$\bigcdot$} & 5.5 \textcolor{green}{$\bigcdot$} & \underline{0.297} & \underline{0.371}\\
& Ours - TaCOS~\cite{yan2025tacos} + NeuralAE~\cite{onzon2021neural} (DF-Grad) & 2.03 \textcolor{green}{$\bigcdot$} & 1.37 \textcolor{green}{$\bigcdot$} & 9.4 \textcolor{green}{$\bigcdot$} & 13.52$\times$6.76 \textcolor{green}{$\bigcdot$} & 6.6 \textcolor{green}{$\bigcdot$} & \textbf{0.300} & \textbf{0.386}\\
\hline
\multirow{4}{*}{Motion Blur$\times$4} & FLIR + nuScenes~\cite{caesar2020nuscenes} + NeuralAE~\cite{onzon2021neural} (Gradient-Based) & 0.43 $\bigcdot$ & 1.65 $\bigcdot$ & 3.6 $\bigcdot$ & 6.2$\times$4.65 $\bigcdot$ & 1.55 $\bigcdot$ & \underline{0.271} & \underline{0.367}\\
& Basler + nuScenes~\cite{caesar2020nuscenes} + NeuralAE~\cite{onzon2021neural} (Gradient-Based) & 0.43 $\bigcdot$ & 1.65 $\bigcdot$ & 3.6 $\bigcdot$ & 4.8$\times$3.6 $\bigcdot$ & 3.75 $\bigcdot$ & 0.255 & 0.344 \\
& TaCOS~\cite{yan2025tacos} + AverageAE~\cite{ARM} & 1.66 \textcolor{green}{$\bigcdot$} & 1.65 \textcolor{green}{$\bigcdot$} & 7.43 \textcolor{green}{$\bigcdot$} & 11.26$\times$5.98 \textcolor{green}{$\bigcdot$} & 5.5 \textcolor{green}{$\bigcdot$} & \underline{0.271} & 0.364\\
& Ours - TaCOS~\cite{yan2025tacos} + NeuralAE~\cite{onzon2021neural} (DF-Grad) & 2.43 \textcolor{green}{$\bigcdot$} & 1.7 \textcolor{green}{$\bigcdot$} & 9.76 \textcolor{green}{$\bigcdot$} & 13.52$\times$6.76 \textcolor{green}{$\bigcdot$} & 6.6 \textcolor{green}{$\bigcdot$} & \textbf{0.303} & \textbf{0.384}\\
\hline
\end{tabular}
}
\vspace{-0.05in}
\label{tab:results}
\end{table*}

\section{Joint Optimisation}\label{sec:Optimization}
\hspace*{\parindent} Fig.~\ref{fig:method_overview} illustrates our joint optimisation of camera hardware, ACC algorithms, and perception tasks. At each training step, the ACC algorithm predicts dynamic parameters from the image from the previous step, which are then used to adjust the image rendered by the camera at the current step. The adjusted image is fed into the perception model for evaluation. The perception task provides a fitness function to update camera hardware and a loss function to update the perception model. Our proposed DF-Grad method then combines the task’s fitness and loss functions to update the ACC network. Optimisation details are given below.

\smallskip \noindent \textbf{Camera Hardware}\hspace{2pt}
We optimise the camera hardware using a GA, jointly with the ACC and perception models. GA is well-suited for this problem because it supports both continuous and discrete camera parameters and is compatible with non-differentiable simulations. This allows us to optimise all hardware parameters supported by the simulation. For example, our first experiment uses the CARLA simulator~\cite{dosovitskiy2017carla}, which supports parameters including placement (location and orientation), optics (focal length and aperture), image sensor (width, height, and pixel count), and multi-camera configurations (number of cameras and their poses). Other parameters can be included depending on the simulator. Further implementation details of the GA are provided in Sec.~\ref{sec:Exp} and the supplementary material.

Following \cite{yan2025tacos}, we adopt the “quantised continuous variables” technique introduced in~\cite{cote2023differentiable} for categorical parameters with internal dependencies. For instance, when selecting an image sensor from a catalogue, the optimiser operates over internal properties such as pixel size and dimensions, by treating them as continuous variables and then replacing them with their nearest catalogue values.

The fitness function guiding this optimisation is defined using the task performance metric.

\smallskip \noindent \textbf{Adaptive Control Neural Network}\hspace{2pt}
The ACC network is trained using our DF-Grad method. Its primary loss is the task loss ($l_{\text{task}}$). To address the non-differentiable rendering of motion blur effects from varying exposure times, we draw inspiration from~\cite{tomasi2021learned} and apply perturbation values $p_e$ and $p_g$ to adjust the predicted exposure time and gain during training. The perturbations are optimised using the GA in the perturbation mode of Fig.~\ref{fig:method_overview}, enabling the ACC network to compensate for performance drops caused by non-differentiably rendered effects. At each training step, $n$ images are rendered using perturbed settings $\tilde{E}$ and $\tilde{G}$, where $\tilde{E} = E_{\text{pred}} + p_e$, $\tilde{G} = G_{\text{pred}} + p_g$, and $n$ is the GA population size. The perturbations that yield the best-performing image are selected and used to compute the loss:
\begin{equation}\label{eqn:GA_loss}
l_{GA} = mean(l_1(\tilde{E}, E_{\text{pred}}),\; l_1(\tilde{G}, G_{\text{pred}})).
\end{equation}

Perturbations are updated at the same rate as the ACC and perception models, but $l_{\text{GA}}$ is only applied every $i$ steps. This allows the GA to find new perturbations when illumination changes. In our experiments, we set $n = 4$ and $i = 4$, with further details in the supplementary material.

In summary, the ACC network is updated with the proposed DF-Grad method using the summed loss of $l_{task}$ and $l_{GA}$. Based on empirical findings (see supplementary material), we weight these terms by 7 for $l_{task}$ and 1 for $l_{GA}$ when computing the summed loss.

\smallskip \noindent \textbf{Perception Task}\hspace{2pt} We focus on deep learning-based perception tasks, which are optimised using gradient-based methods alongside the camera hardware and ACC network. The task loss updates both the task model and the ACC network, while a performance metric derived from the task guides optimisation of the camera hardware and perturbations.

In training, the ACC and perception models are updated every step simultaneously, while camera hardware is updated every $k$ steps. The fitness function is accumulated at each step, and its mean over $k$ steps is used to update the camera hardware. This ensures that each design is evaluated across varying illumination conditions. The choice of $k$ depends on the perception task and the rate of illumination change. In our experiments, we set $k = 45$, balancing wide scenario coverage with reasonable optimisation time.

%% file: sec/6_Experiment.tex
\section{Experiments}\label{sec:Exp}
\hspace*{\parindent} We validate our method on a camera design task for object detection in autonomous driving using both synthetic and real-world images. Experiments span multiple configurations, compared against adaptive control and camera design baselines, and analysed through ablation studies.

\subsection{Implementations}\label{sec:exp_setup}
\smallskip \noindent \textbf{Noise and Brightness Calibration}\hspace{2pt}
We calibrate brightness and noise using a Basler Dart DaA1280-54uc camera~\cite{Basler2024}, with an exposure time of 5 ms, a gain of 10 dB, and an aperture of f/16. These settings ensure unsaturated images under daylight with a large DoF. Noise calibration follows~\cite{wang2021multiplexed, yan2025tacos}, and brightness is calibrated by matching intensity histograms of physically captured and images captured by the simulator or from the dataset.

\smallskip \noindent \textbf{Perception Task}\hspace{2pt}
The perception task is object detection with Faster R-CNN~\cite{ren2015faster}. We use a ResNet-50~\cite{he2016deep} backbone and pretrain the model on ImageNet~\cite{deng2009imagenet}. Training uses the AdamW optimiser~\cite{loshchilov2017decoupled} with batch size 1, learning rate $1\cdot10^{-4}$, and weight decay $1\cdot10^{-6}$.

\smallskip \noindent \textbf{Camera Hardware Optimisation}\hspace{2pt}
Camera hardware parameters are optimised using a GA with 10 candidates per generation over 35 generations, totalling 350 designs. Fitness is defined as the mean Average Precision (mAP) averaged over intersection over union (IoU) thresholds from 0.5 to 0.95. When optimising the camera's FoV, we add the ratio of true positives at 0.5 IoU to all visible objects in a $180^\circ$ horizontal FoV as a secondary metric. This metric avoids overfitting to narrow FoVs that may inflate mAP. We assign a weight of 1 to the mAP term and 1.1 to the true positive (TP) ratio term, based on empirical tuning. 

\smallskip \noindent \textbf{Adaptive Control Network Optimisation}\hspace{2pt} The ACC model is trained using Adam~\cite{kingma2015adam} with a learning rate of $1\cdot10^{-5}$ and a batch size of 1. The fitness function for optimising perturbations is the AP at an IoU threshold of 0.5, using a single threshold to reduce training time.

The supplementary material provides further details on GA implementation for perturbations and camera hardware.

\subsection{Comparison}\label{sec:compare}
\hspace*{\parindent} We compare our method against existing approaches. These include using TaCOS~\cite{yan2025tacos} to jointly optimise camera hardware and task, combined with a non-trainable average-based AE algorithm (AverageAE~\cite{ARM}), as well as using the neural auto-exposure method (NeuralAE~\cite{onzon2021neural}) with off-the-shelf cameras, the Basler Dart DaA1280-54uc~\cite{Basler2024} and FLIR Flea3~\cite{FLIR2017} cameras, using human-designed placements from nuScenes~\cite{caesar2020nuscenes}. For fairness, TaCOS is run with the same GA configuration as our method (10 candidates per generation, 35 generations). The RL-based DISeR~\cite{klinghoffer2023diser} is excluded, as it was developed and demonstrated for single-frame tasks~\cite{klinghoffer2023diser, yan2025tacos}. In our work, each camera configuration needs to be evaluated over multiple frames to determine its performance over changing light conditions, resulting in impractically long optimisation times.

AverageAE adjusts exposure using a scaling factor $0.5M_{\text{white}}/I_{mean}$, where $I_{mean}$ is the mean image intensity, driving the intensity toward half of $M_{white}$. We scale exposure time and gain by the square root of this factor, following~(\ref{eqn:image_formation}). The NeuralAE implementation used in this work follows the description in Sec.~\ref{sec:AEC}, which is slightly modified from the original version. We use exposure settings from the previous timestep as additional inputs to the neural network and let it control both exposure time and gain.

We first evaluate all methods under a standard scenario with calibrated image noise and motion blur. Autonomous platforms can operate in challenging scenarios, including darker scenes, stronger image noise, and faster motion, we additionaly test under increased noise levels for lower SNRs, representing a reduced average illumination (less signal) or stronger sensor noise, and under stronger motion blur to represent faster motion on average. In all scenarios, illumination changes dynamically between maximum brightness and darkness, and the vehicles' speed changes constantly between 0 and maximum.
\begin{table}
\centering
\caption{
Experiment results with Waymo Open dataset. Again, we compare our method against existing methods on the ACC algorithm and camera design under different scenarios. The results show that our method consistently outperforms existing methods, especially under challenging conditions.
}
\resizebox{\columnwidth}{!}{%
\begin{tabular}{ccccc}
\hline
\multirow{3}{*}{Scenario} & \multirow{3}{*}{Method} & \multicolumn{2}{c}{Camera Parameters} & Performance \\
\cline{3-5}
&  & Sensor Size & Pixel Size &  Object Detect. \\
& & $w\times h$  (mm) & $p$ ($\mu$m) & mAP $\uparrow$ \\
\hline
\multirow{3}{*}{Calibrated} & Basler + NeuralAE~\cite{onzon2021neural} & 4.8$\times$3.6 $\bigcdot$ & 3.75 $\bigcdot$ & \underline{0.316} \\
& TaCOS~\cite{yan2025tacos} + AverageAE~\cite{ARM} & 4.8$\times$3.6 \textcolor{green}{$\bigcdot$} & 3.75 \textcolor{green}{$\bigcdot$} & 0.297 \\
& Ours & 5.38$\times$3.02 \textcolor{green}{$\bigcdot$} & 2.8 \textcolor{green}{$\bigcdot$} & \textbf{0.331}\\
\hline
\multirow{3}{*}{Noise$\times$10} & Basler + NeuralAE~\cite{onzon2021neural} & 4.8$\times$3.6 $\bigcdot$ & 3.75 $\bigcdot$ & 0.303 \\
& TaCOS~\cite{yan2025tacos} + AverageAE~\cite{ARM} & 7.31 $\times$ 5.58 \textcolor{green}{$\bigcdot$} & 4.5 \textcolor{green}{$\bigcdot$} & \underline{0.308} \\
& Ours & 7.31$\times$5.58 \textcolor{green}{$\bigcdot$} & 4.5 \textcolor{green}{$\bigcdot$} & \textbf{0.322} \\
\hline
\multirow{3}{*}{Noise$\times$20} & Basler + NeuralAE~\cite{onzon2021neural} & 4.8$\times$3.6 $\bigcdot$ & 3.75 $\bigcdot$ & 0.272 \\
& TaCOS~\cite{yan2025tacos} + AverageAE~\cite{ARM} & 6.14$\times$4.92 \textcolor{green}{$\bigcdot$} & 4.8 \textcolor{green}{$\bigcdot$} & \underline{0.289} \\
& Ours & 7.31$\times$5.58 \textcolor{green}{$\bigcdot$} & 4.5 \textcolor{green}{$\bigcdot$} & \textbf{0.319} \\
\hline
\multirow{3}{*}{Motion Blur$\times$2} & Basler + NeuralAE~\cite{onzon2021neural} & 4.8$\times$3.6 $\bigcdot$ & 3.75 $\bigcdot$ & \underline{0.304} \\
& TaCOS~\cite{yan2025tacos} + AverageAE~\cite{ARM} & 11.25$\times$7.03 \textcolor{green}{$\bigcdot$} &  5.86 \textcolor{green}{$\bigcdot$} &  0.293 \\
& Ours & 6.62$\times$4.14 \textcolor{green}{$\bigcdot$} & 3.45\textcolor{green}{$\bigcdot$} & \textbf{0.316}\\
\hline
\multirow{3}{*}{Motion Blur$\times$4} & Basler + NeuralAE~\cite{onzon2021neural} & 4.8$\times$3.6 $\bigcdot$ & 3.75 $\bigcdot$ & 0.269  \\
& TaCOS~\cite{yan2025tacos} + AverageAE~\cite{ARM} & 8.45$\times$6.76 \textcolor{green}{$\bigcdot$} & 6.6 \textcolor{green}{$\bigcdot$} & \underline{0.272} \\
& Ours & 11.25$\times$7.03 \textcolor{green}{$\bigcdot$} & 5.86 \textcolor{green}{$\bigcdot$} & \textbf{0.317} \\
\hline
\end{tabular}
}
\vspace{-0.05in}
\label{tab:results_real}
\end{table}

\begin{figure*}
        \centering
        \includegraphics[width=\linewidth]{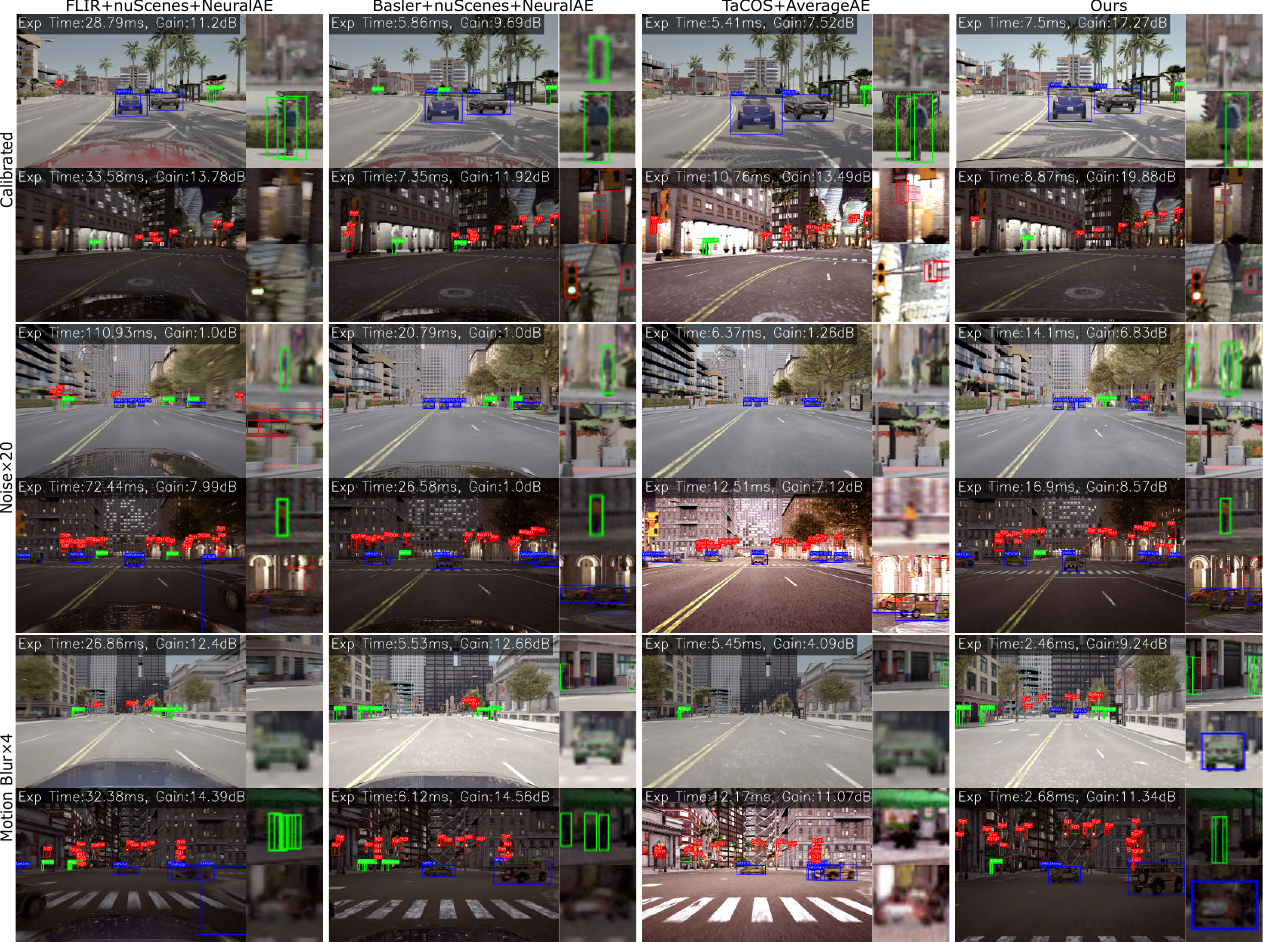}
        \caption{Qualitative results using CARLA simulator. Comparison of the proposed method with baseline approaches that jointly optimise NeuralAE and the object detector using human-designed cameras, as well as with methods that jointly optimise camera hardware and the object detector using the non-trainable AverageAE algorithm, across different design scenarios. Our results demonstrate that the jointly optimised camera and ACC algorithm from our method consistently produces sharp images with high effective object resolution and reduced motion blur. This leads to improved task performance, particularly in detecting small and distant objects.}
\label{fig:result}
\end{figure*}

\subsection{Synthetic Images}\label{sec:results}
\hspace*{\parindent} Synthetic images are collected in the CARLA simulator~\cite{dosovitskiy2017carla}, with a camera mounted on an autonomous vehicle. Illumination is varied across daytime, nighttime, abrupt transitions between day and night, and gradual transitions simulating continuous day-night cycles. Motion blur is rendered by the simulator based on camera and object motion as well as exposure time. During training, each camera configuration captures 45 frames, producing 15750 images in total. Testing uses 1000 images collected on a separate urban map. The dataset contains three object categories: vehicles, walkers, and traffic signs. We compare our method to all baseline methods discussed in Sec.~\ref{sec:compare}.

\smallskip \noindent \textbf{Design Space}\hspace{2pt} The design space includes the camera’s forward position ($x \in [0\;\text{m}, 2.4\;\text{m}]$) and height ($z \in [1.3\;\text{m}, 1.7\;\text{m}]$), constrained by vehicle geometry. The origin is at the vehicle centre at ground level. The camera is fixed along the midline in the horizontal axis to minimise blur during turning. Additional optimised parameters include focal length ($f \in [1\;\text{mm}, 10\;\text{mm}]$) and image sensor characteristics (width $w$, height $h$, pixel size $p$), drawn from a catalogue of 43 commercial CMOS sensors~\cite{yan2025tacos} and optimised as categorical variables using the quantised continuous method as explained in Sec.~\ref{sec:Optimization}. Based on our image brightness calibration and the settings of the physical camera, we let the ACC network predict an exposure time between 0.1 ms to 50 ms, and a gain between 1 dB to 30 dB. Perturbations are restricted to $\pm 5\%$ of the parameter ranges to maintain temporal consistency and avoid flickering.

\subsection{Real-World Images}\label{sec:result_real} 
\hspace*{\parindent}To further evaluate our method, we use physically captured images from the Waymo Open Perception Dataset~\cite{sun2020waymo}, specifically the front-left camera.  The images are captured under varying illumination. We select around 2500 images for training, and 350 camera hardware configurations are optimised over 15750 steps, matching the synthetic setup. An additional 550 images are used for testing. The object classes include vehicles, walkers, and cyclists.

To emulate the image formation process, intensities are rescaled to the level of our calibrated images using the dataset's exposure time metadata, treating the adjusted images as $I_0$ in~(\ref{eqn:image_scale}). We restrict the design space to sensors with pixel counts smaller than the pixel count of the dataset ($1920\times1280$). The rescaled images are then downsized with bilinear interpolation based on the current camera configuration. Noise and motion blur in the original image are ignored, and we add them to the images based on camera parameters. Noise is added using the same affine model as in the synthetic setup, and motion blur is simulated by approximating the exposure integral via optical flow. Optical flow is estimated with Dense Inverse Search~\cite{kroeger2016fast} between consecutive frames. As the effects of resolution, noise, and motion blur are emulated through post-processing rather than captured with physical hardware, the experiment is constrained by the fidelity of the post-processing methods.

We compare our method with all baselines in Sec.~\ref{sec:compare} except the FLIR camera with the NeuralAE method, as this camera exceeds the pixel count of the original images.

\smallskip \noindent \textbf{Design Space}\hspace{2pt} Since FoV and placement are fixed in the dataset, the design space is limited to image sensor parameters ($w$, $h$, $p$). These are selected from the same catalogue as in the synthetic experiment but restricted to sensors with pixel counts smaller than the dataset resolution ($1920\times1280$), resulting in 17 candidates. Dynamic parameters and their perturbations are optimised over the same ranges as in the synthetic experiment. The same object detection model is jointly optimised for this dataset.

\begin{table*}
\centering
\caption{
Ablation study on learning the ACC network without the GA supervision, with fixed perturbation values, and freezing the object detection network when optimising the camera hardware and ACC network.
}
\resizebox{2\columnwidth}{!}{%
\begin{tabular}{cccccccc}
\hline
\multirow{3}{*}{Method} & \multicolumn{5}{c}{Camera Parameters} & \multicolumn{2}{c}{Performance} \\
\cline{2-8}
& Forward Position & Height & Focal Length & Sensor Size & Pixel Size &  Object Detect. & True Positive Ratio\\
& x ($m$) & z ($m$) & $f$ (mm) & $w\times h$  (mm) & $p$ ($\mu$m) & mAP $\uparrow$ & $\text{TP}/\text{All@}180^{\circ}\uparrow$\\
\hline
No GA Component - TaCOS~\cite{yan2025tacos} + NeuralAE~\cite{onzon2021neural} (Gradient-Based) & 2.43 \textcolor{green}{$\bigcdot$} & 1.65 \textcolor{green}{$\bigcdot$} & 8.75 \textcolor{green}{$\bigcdot$} & 11.34$\times$7.13 \textcolor{green}{$\bigcdot$} & 5.86 \textcolor{green}{$\bigcdot$} & 0.310 & 0.379\\
Fixed Perturbations - TaCOS~\cite{yan2025tacos} + NeuralAE~\cite{onzon2021neural} (Fixed Pertrub) & 1.68 \textcolor{green}{$\bigcdot$} & 1.25 \textcolor{green}{$\bigcdot$} & 8.14 \textcolor{green}{$\bigcdot$} & 11.25$\times$7.03 \textcolor{green}{$\bigcdot$} & 5.86 \textcolor{green}{$\bigcdot$} & 0.302 & 0.372\\
Frozen Object Detector - TaCOS~\cite{yan2025tacos} + NeuralAE~\cite{onzon2021neural} (DF-Grad) & 1.31 \textcolor{green}{$\bigcdot$} & 1.3 \textcolor{green}{$\bigcdot$} & 5.94 \textcolor{green}{$\bigcdot$} & 8.66$\times$4.34 \textcolor{green}{$\bigcdot$} & 2.25 \textcolor{green}{$\bigcdot$} & 0.233 & 0.367\\
Ours - TaCOS~\cite{yan2025tacos} + NeuralAE~\cite{onzon2021neural} (DF-Grad) & 1.35 \textcolor{green}{$\bigcdot$} & 1.3 \textcolor{green}{$\bigcdot$} & 5.19 \textcolor{green}{$\bigcdot$} & 7.37$\times$4.92 \textcolor{green}{$\bigcdot$} & 2.4 \textcolor{green}{$\bigcdot$} & \textbf{0.325} & \textbf{0.390}\\
\hline
\end{tabular}
}
\vspace{-0.05in}
\label{tab:ablation}
\end{table*}

\subsection{Results}
\hspace*{\parindent} Tab.~\ref{tab:results} and Tab.~\ref{tab:results_real} report quantitative results for experiments on synthetic and real-world data. For synthetic images, performance is measured using mAP@0.5-0.95 IoU and the TP ratio; for real-world data, only mAP is reported as FoV is fixed. Qualitative results from the synthetic data are displayed in Fig.~\ref{fig:result}, and qualitative results from the real-world data are presented in the supplementary material.

\smallskip \noindent \textbf{Synthetic Images}\hspace{2pt} Under calibrated conditions with high SNR and mild motion blur, our method achieves slightly better performance than existing approaches. In more challenging scenarios, it significantly outperforms all baselines.

In low-SNR scenarios with stronger noise, NeuralAE increases exposure time and reduces gain to reduce noise, following ~(\ref{eqn:image_formation}) and~(\ref{eqn:noise_generlization}). This amplifies motion blur, degrading task performance. The effect is most severe with the FLIR camera, which requires longer exposure due to its small pixels. In contrast, our method effectively balances exposure time and gain through the DF-Grad method while adapting the hardware design. Specifically, it selects larger pixels under high-noise conditions to improve SNR, and smaller pixels in high-SNR settings to maximise resolution. The method using TaCOS combined with AverageAE fails to adapt to noise, as AverageAE adjusts the dynamic parameters only based on mean intensity. Meanwhile, TaCOS compensates for the low SNR by using large pixels.

With stronger motion blur, NeuralAE without GA retains exposure settings close to the calibrated case, as it cannot learn non-differentiable effects. Our method uses GA to infer the impact of motion blur on task performance, shortening exposure time and increasing gain to mitigate blur while maintaining brightness. Both our method and TaCOS also select larger pixels to reduce exposure time and gain, thereby lowering motion blur effects and noise.

Finally, we observe variation in camera placement across scenarios for both our method and TaCOS due to limited constraints from the detection task, which only requires an unobstructed view. Both methods also maintain appropriate FoVs by correlating sensor size and focal length.

\smallskip \noindent \textbf{Real-World Images}\hspace{2pt} Experiments on real-world images show consistent trends. Our method outperforms baselines across calibrated, noisy, and motion-blurred conditions. Despite a smaller design space, it successfully balances SNR, resolution, and motion blur by optimising the pixel sizes and exposure parameters, while baseline methods cannot. NeuralAE without GA tends to maximise exposure to suppress noise, ignoring motion blur, and TaCOS with AverageAE lacks the flexibility to adjust dynamic parameters under challenging scenarios.

\subsection{Ablation Study}
\hspace*{\parindent} We conduct ablations to assess the effectiveness of key components in our method using the CARLA simulator under the calibrated scenario. First, we remove the GA component and train the ACC network following the original NeuralAE procedure~\cite{onzon2021neural}. Next, we replace GA-optimised perturbations with fixed ones set to half of their ranges (2.5\% of their maximum values). Using the same population size and frequency ($n=4,\; i=4$), we generate four perturbation combinations: $(E_{\text{pred}} \pm p_e,\; G_{\text{pred}} \pm p_g)$, selecting the one with the highest task performance to compute the GA supervision loss in~\ref{eqn:GA_loss}.
Lastly, we evaluate the effect of freezing the object detection network during training. In this setup, the detector is pretrained on 15750 images from 350 random camera configurations, matching the data and camera configurations used in the joint training schemes.

Without GA, the method selects cameras with larger pixel sizes to reduce motion blur, but this lowers effective resolution and degrades detection accuracy. We observe slightly lower performance than the FLIR camera with NeuralAE baseline, likely due to the limited number of generations. Since the camera design evolves throughout training, the perception network must adapt continuously, and insufficient data per configuration leads to suboptimal performance. This can be mitigated by using more generations or perception models pretrained on the dataset to stabilise early training as in~\cite{yan2025tacos}. Our DF-Grad method offsets this drop by learning a more effective ACC network. Nevertheless, even without GA, the optimised camera outperforms FLIR in high motion blur scenarios such as vehicle turns.

With fixed perturbations, the ACC network tends to predict shorter exposures and higher gains, leading to larger pixel sizes and lower resolution. Although this helps reduce motion blur, it fails to balance SNR. In addition, this method fails to effectively update the perturbations based on the ACC predictions, leading to less desirable dynamic parameters and therefore reducing the performance. In contrast, our GA-based approach dynamically adapts to scene context, resulting in better overall performance.

Freezing the detection network causes a significant performance drop, as it cannot adapt to new camera parameters. This highlights the need for joint optimisation of both camera design and perception.

%% file: sec/7_Conclusion.tex
\section{Conclusion} \label{sec:Conclusion}
\hspace*{\parindent} We present a novel end-to-end approach that jointly optimises camera hardware, ACC algorithms, and perception tasks. In particular, we propose DF-Grad, a method combining derivative-free and gradient-based optimisers to train neural network-based ACC algorithms, effectively handling the non-differentiable nature of image formation. The method supports the optimisation of continuous and discrete static camera parameters, as well as dynamic camera parameters. This framework can be directly generalised to other perception tasks and ACC algorithms that control other dynamic camera parameters, such as aperture diameter and focus distance, representing an important step toward the automated design of imaging systems that prioritise task performance in dynamic environments. In future work, we aim to extend this approach to adverse weather conditions with physically realistic image formation methods, and to more complex camera design systems involving unconventional cameras and multi-camera designs.

\smallskip \noindent \textbf{Acknowledgements}\hspace{2pt} This research was supported in part through the NVIDIA Academic Grant Program.

%% file: supplementary.tex

\begin{figure}
      \centering
      \includegraphics[width=\linewidth]{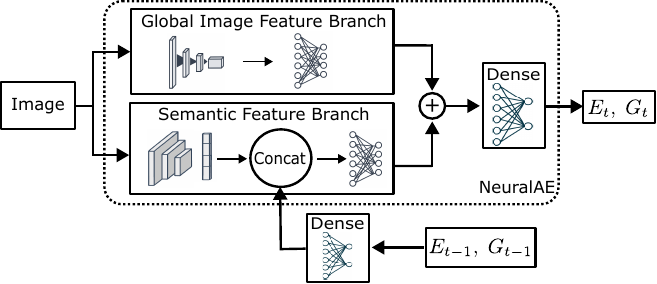}
      \caption{Adaptive camera control algorithm. We adopt the architecture from NeuralAE~\cite{onzon2021neural} as the main architecture for the ACC algorithm. We modify NeuralAE by using a single camera instead of two cameras, and by concatenating features from the predicted dynamic parameters at the previous step to the extracted features in the semantic feature branch for temporal consistency. Finally, we allow it to predict both exposure time and gain, rather than a single exposure value as in its original version. Figure is adapted and modified from~\cite{onzon2021neural}.}
      \label{fig:ACC_model}
\end{figure}

\section{Architecture of Adaptive Camera Control Model}
We adopt and modify the NeuralAE model from~\cite{onzon2021neural} as the adaptive camera control (ACC) algorithm. An illustration of the model architecture is shown in Fig.~\ref{fig:ACC_model}.

The model retains both the global image feature branch and the semantic feature branch from the original implementation. The global image feature branch uses a 3-layer 1D convolutional neural network to extract features from multi-scale histograms of the images, followed by two linear layers to obtain dense features.

The semantic feature branch reuses the ResNet~\cite{he2016deep} feature extractor from the downstream tasks, taking the conv2 layer output as input. This feature is compressed with a 2D convolution and split into three compressed feature maps (CFMs). Average pooling is applied to the first CFM, while max pooling is applied to the last two. The original compressed feature (before splitting) also undergoes average pooling. These pooled features from the CFMs and the compressed feature are concatenated. We modify this step by also concatenating a 32-dimensional feature map derived from the dynamic parameters predicted at the previous timestep. This feature map is obtained by expanding the two input parameters through two linear layers. The concatenated feature then passes through two linear layers to produce a dense feature of the same size as the global image feature branch output. To accommodate the additional features, the input size of these linear layers is increased by 32 (original: 784, modified: 816).

Finally, the outputs of the two branches are summed, and a linear layer is applied to reduce the summed feature to match the number of required dynamic parameters. Unlike the original implementation in~\cite{onzon2021neural}, which outputs only a single exposure value, our modified model predicts both exposure time and gain.

Further architectural details can be found in~\cite{onzon2021neural}. We maintain the same number of parameters for each layer as in the original, with the only modifications being the concatenation of previous dynamic parameter features and the expanded output size.

\section{Details on Optimisation}
\hspace*{\parindent} We apply a genetic algorithm (GA)~\cite{holland1992adaptation} to optimise the camera hardware, and use the proposed DF-Grad method to train the ACC algorithm. The implementations of these methods are detailed in this section.

\subsection{Camera Hardware} 
\hspace*{\parindent} We use a population size of 10 over 35 generations, with 50\% of the population generating offspring through crossover and mutation. A uniform crossover process is applied, where each parameter of the offspring is randomly selected from one of the parents. The mutation process multiplies each parameter by a random factor between 0.8 and 1.2. Initial camera parameters are randomly sampled from their respective feasible ranges. These hyperparameters and implementation choices follow the recommendations of~\cite{yan2025tacos}, which optimises a similar number of parameters in its experiments.

\subsection{Adaptive Camera Control Algorithm}
\hspace*{\parindent} We determine the weights of GA loss and task loss when updating the ACC model, the population size used to optimise perturbations for dynamic parameters, and the frequency of applying the GA loss based on experimental results. Preliminary experiments are first conducted to identify suitable value ranges for these hyperparameters. We then select the final values through empirical tuning based on performance. The results and analysis from the empirical value selection in the chosen ranges are presented in this section.

\begin{table*}[http]
\centering
\caption{
Camera designs and object detection performance for task loss weights of 5, 7, and 9 used during ACC algorithm training. The results indicate that a weight of 7 yields the best task performance.
}
\resizebox{2\columnwidth}{!}{%
\begin{tabular}{cccccccc}
\hline
\multirow{3}{*}{Weight} & \multicolumn{5}{c}{Camera Parameters} & \multicolumn{2}{c}{Performance} \\
\cline{2-8}
& Forward Position & Height & Focal Length & Sensor Size & Pixel Size &  Object Detect. & True Positive Ratio\\
& x ($m$) & z ($m$) & $f$ (mm) & $w\times h$  (mm) & $p$ ($\mu$m) & mAP $\uparrow$ & $\text{TP}/\text{All@}180^{\circ}\uparrow$\\
\hline
5 & 1.34 \textcolor{green}{$\bigcdot$} & 1.35 \textcolor{green}{$\bigcdot$} & 4.91 \textcolor{green}{$\bigcdot$} & 5.70$\times$4.28 \textcolor{green}{$\bigcdot$} & 2.2 \textcolor{green}{$\bigcdot$} & 0.293 & 0.366\\
7 (Proposed) & 1.35 \textcolor{green}{$\bigcdot$} & 1.3 \textcolor{green}{$\bigcdot$} & 5.19 \textcolor{green}{$\bigcdot$} & 7.37$\times$4.92 \textcolor{green}{$\bigcdot$} & 2.4 \textcolor{green}{$\bigcdot$} & \textbf{0.325} & \textbf{0.390}\\
9 & 1.31 \textcolor{green}{$\bigcdot$} & 1.30 \textcolor{green}{$\bigcdot$} & 5.76 \textcolor{green}{$\bigcdot$} & 6.77$\times$5.66 \textcolor{green}{$\bigcdot$} & 2.74 \textcolor{green}{$\bigcdot$} & 0.296 & 0.361\\
\hline
\end{tabular}
}
\vspace{-0.05in}
\label{tab:weight}
\end{table*}
\begin{table*}[!ht]
\centering
\caption{
Camera designs and object detection performance for population sizes of 3, 4, and 5 used in GA optimisation of dynamic parameter perturbations. The results show that a population size of 4 achieves the best task performance.
}
\resizebox{2\columnwidth}{!}{%
\begin{tabular}{cccccccc}
\hline
\multirow{3}{*}{Population Size} & \multicolumn{5}{c}{Camera Parameters} & \multicolumn{2}{c}{Performance} \\
\cline{2-8}
& Forward Position & Height & Focal Length & Sensor Size & Pixel Size &  Object Detect. & True Positive Ratio\\
& x ($m$) & z ($m$) & $f$ (mm) & $w\times h$  (mm) & $p$ ($\mu$m) & mAP $\uparrow$ & $\text{TP}/\text{All@}180^{\circ}\uparrow$\\
\hline
3 & 2.2 \textcolor{green}{$\bigcdot$} & 1.57 \textcolor{green}{$\bigcdot$} & 6.09 \textcolor{green}{$\bigcdot$} & 8.66$\times$4.34 \textcolor{green}{$\bigcdot$} & 2.25 \textcolor{green}{$\bigcdot$} & 0.296 & 0.388\\
4 (Proposed) & 1.35 \textcolor{green}{$\bigcdot$} & 1.3 \textcolor{green}{$\bigcdot$} & 5.19 \textcolor{green}{$\bigcdot$} & 7.37$\times$4.92 \textcolor{green}{$\bigcdot$} & 2.4 \textcolor{green}{$\bigcdot$} & \textbf{0.325} & \textbf{0.390}\\
5 & 1.80 \textcolor{green}{$\bigcdot$} & 1.70 \textcolor{green}{$\bigcdot$} & 9.81 \textcolor{green}{$\bigcdot$} & 14.13$\times$7.45 \textcolor{green}{$\bigcdot$} & 3.45 \textcolor{green}{$\bigcdot$} & 0.306 & 0.381\\
\hline
\end{tabular}
}
\vspace{-0.05in}
\label{tab:pop_size}
\end{table*}
\begin{table*}[!ht]
\centering
\caption{
Camera designs and object detection performance when updating the ACC algorithm with GA loss every 3, 4, and 5 steps. The results show that an update frequency of 4 achieves the best task performance.
}
\resizebox{2\columnwidth}{!}{%
\begin{tabular}{cccccccc}
\hline
\multirow{3}{*}{Update Frequency} & \multicolumn{5}{c}{Camera Parameters} & \multicolumn{2}{c}{Performance} \\
\cline{2-8}
& Forward Position & Height & Focal Length & Sensor Size & Pixel Size &  Object Detect. & True Positive Ratio\\
& x ($m$) & z ($m$) & $f$ (mm) & $w\times h$  (mm) & $p$ ($\mu$m) & mAP $\uparrow$ & $\text{TP}/\text{All@}180^{\circ}\uparrow$\\
\hline
3 & 1.70 \textcolor{green}{$\bigcdot$} & 1.70 \textcolor{green}{$\bigcdot$} & 5.94 \textcolor{green}{$\bigcdot$} & 8.66$\times$4.34 \textcolor{green}{$\bigcdot$} & 2.25 \textcolor{green}{$\bigcdot$} & 0.301 & 0.379\\
4 (Proposed) & 1.35 \textcolor{green}{$\bigcdot$} & 1.3 \textcolor{green}{$\bigcdot$} & 5.19 \textcolor{green}{$\bigcdot$} & 7.37$\times$4.92 \textcolor{green}{$\bigcdot$} & 2.4 \textcolor{green}{$\bigcdot$} & \textbf{0.325} & \textbf{0.390}\\
5 & 2.01 \textcolor{green}{$\bigcdot$} & 1.70 \textcolor{green}{$\bigcdot$} & 4.51 \textcolor{green}{$\bigcdot$} & 5.70$\times$4.28 \textcolor{green}{$\bigcdot$} & 2.2 \textcolor{green}{$\bigcdot$} & 0.320 & 0.381\\
\hline
\end{tabular}
}
\vspace{-0.05in}
\label{tab:update_freq}
\end{table*}
\begin{figure*}[!ht]
      \centering
      \includegraphics[width=\linewidth]{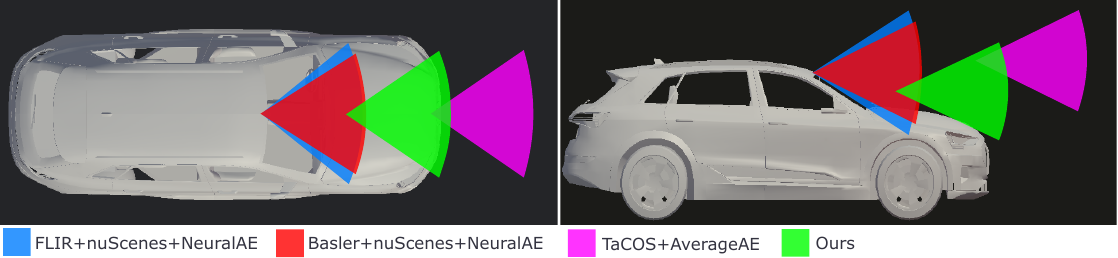}
      \caption{Visualisation of the camera FoVs and placements for our designed camera, the FLIR/Basler cameras using the nuScenes placement, and the camera designed by TaCOS with the AverageAE method.}
      \label{fig:FoV}
\end{figure*}
\begin{figure*}[!ht]
      \centering
      \includegraphics[width=\linewidth]{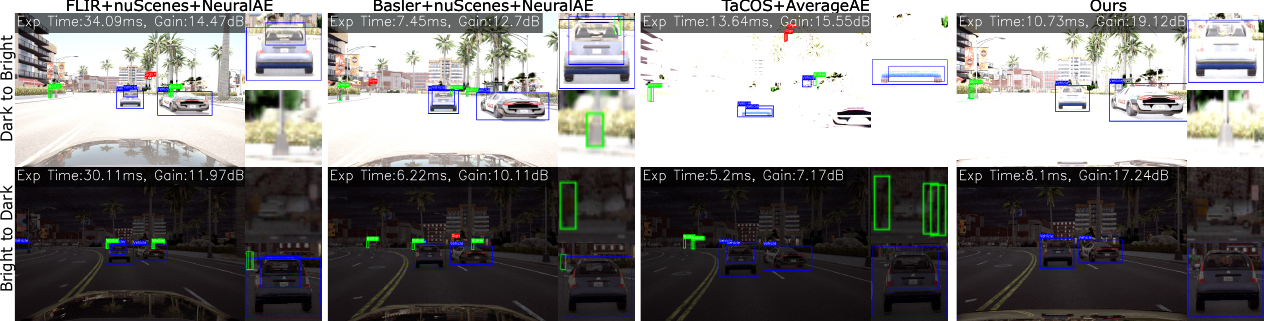}
      \caption{Qualitative results under abrupt illumination change from dark to bright and vice versa. The results shows an increased detection rate with the NeuralAE method, highlighting the advantages of using a learning-based ACC algorithm in handling such transitions.}
      \label{fig:rapid_change}
\end{figure*}
\begin{figure*}[http]
      \centering
      \includegraphics[width=0.99\linewidth]{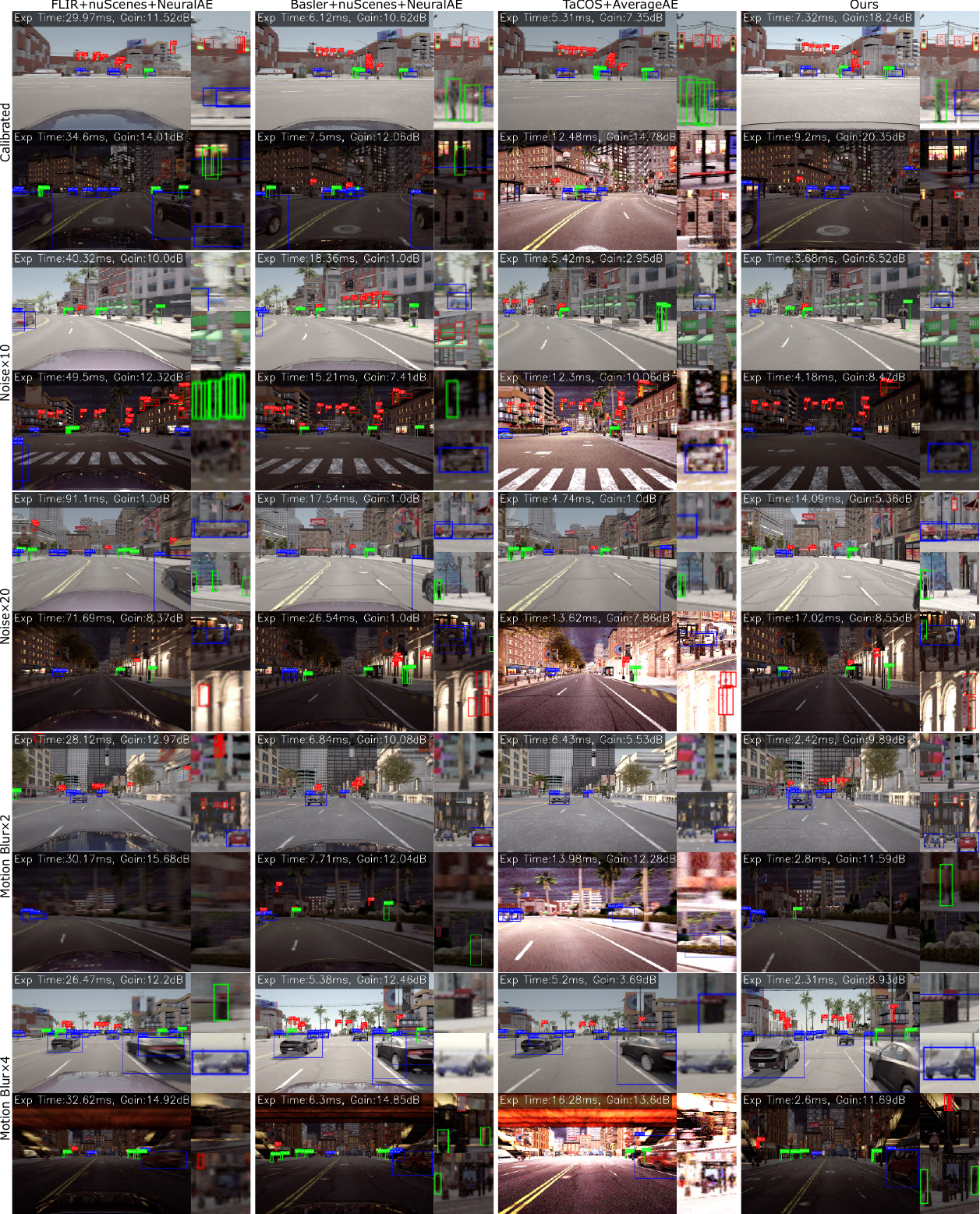}
      \caption{Additional qualitative results on synthetic data comparing our proposed approach with baselines. The results demonstrate that our method more accurately detects small and distant objects, particularly in challenging scenarios with increased noise and motion blur.}
      \label{fig:addition_results}
\end{figure*}

\begin{figure*}[http]
      \centering
      \includegraphics[width=0.91\linewidth]{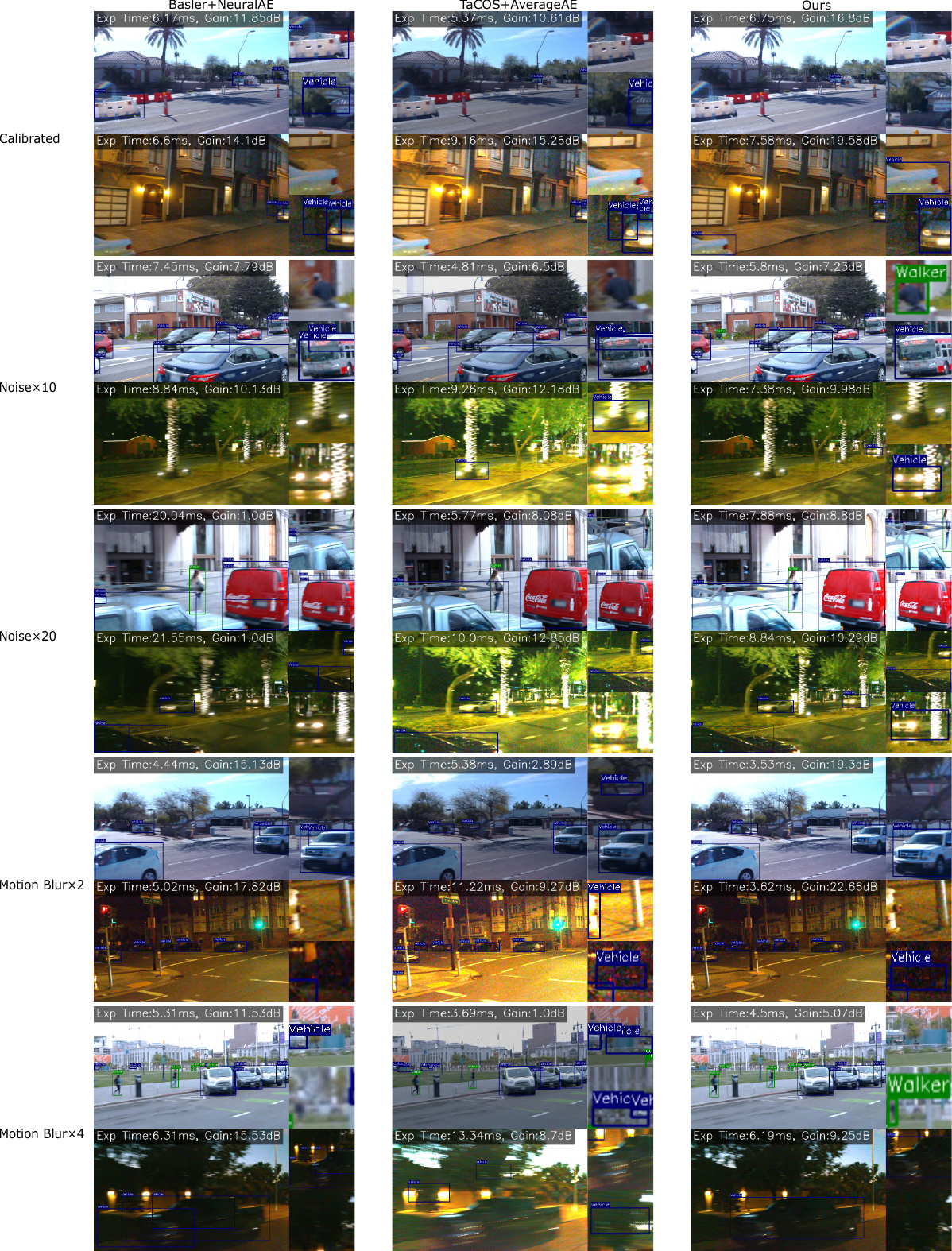}
      \caption{
      Qualitative results on real-world images show that our method consistently outperforms all baselines, further validating its practicality. Competing methods exhibit noticeable false positives, such as the examples on rows 1 and 10, and other examples show a mix of both false positives and false negatives from the baselines.}
      \label{fig:waymo_results}
\end{figure*}

\smallskip \noindent \textbf{Weight of Loss Function} Our DF-Grad optimisation scheme uses a combined task loss and GA loss, as described in Sec. 5 of the main paper, to train the ACC algorithm. We fix the weight of the GA loss at 1, and vary the task loss weight to determine the optimal value. We show the results of using a weight of 5, 7, and 9 in Tab.~\ref{tab:weight}. The results suggest that a weight of 7 best balances the unsupervised learning signal from the task model with the supervision from GA, yielding the best task performance. The results also indicate that the camera hardware design is relatively insensitive to changes in the task loss weight.

\smallskip \noindent \textbf{Population Size} We conduct experiments with varying population sizes for the GA optimisation of exposure perturbations. Results are summarised in Tab.~\ref{tab:pop_size}. A population size of 4 provides the best performance. Smaller population sizes cause the optimiser to converge to local optima due to reduced diversity, while larger population sizes increase convergence time~\cite{yan2025tacos}, potentially requiring a longer update frequency and thus lengthier overall optimisation.

\smallskip \noindent \textbf{Update Frequency} As discussed in the main paper, the GA loss is used to supervise the ACC network every $i$ steps. The update frequency $i$ represents the number of steps allowed for GA to optimise perturbations before applying the supervision. Since illumination conditions can change frequently, applying the GA loss intermittently allows new perturbations to be optimised for each lighting condition. We show the results of using an update frequencies of 3, 4, and 5 in Tab.~\ref{tab:update_freq}. The results indicate that applying GA loss every 4 steps offers a good balance, giving the GA sufficient time to adapt to changing environments while providing consistent supervisory signals to the ACC algorithm.

We note that population size and update frequency are closely related. A larger population size generally requires a larger value of update frequency, as it increases convergence time~\cite{yan2025tacos}. In our experiments, we first select a population size based on optimisation time and early performance with an intuitive update frequency, then refine the update frequency through empirical testing.

\smallskip \noindent \textbf{GA Implementation} We optimise the perturbations of dynamic parameters using a GA. Following~\cite{yan2025tacos}, 50\% of the population generates offspring via uniform crossover. For mutation, we apply a random multiplication factor between 0.9 and 1.1 to each parameter. This tighter range avoids abrupt changes in dynamic settings, which could lead to undesirable visual artefacts such as flickering.

\section{Additional Results on Camera Design}
\hspace*{\parindent} In this section, we present additional results from our camera design experiments. These include illustrations of the cameras’ fields of view (FoV) and placements, performance under rapid illumination changes, and additional qualitative comparisons.

\subsection{Camera FoV and Placement}
\hspace*{\parindent} We illustrate the FoVs and placements of the camera designed by our proposed method, the FLIR~\cite{FLIR2017} and Basler~\cite{Basler2024} cameras (using the nuScenes placement~\cite{caesar2020nuscenes}), and the camera designed by TaCOS~\cite{yan2025tacos} with the AverageAE~\cite{ARM} method in Fig.~\ref{fig:FoV}. These visualisations correspond to the calibrated design scenario.

From the figure, we observe that both our method and TaCOS produce camera designs with moderate FoVs, striking a balance between maximising mAP and maintaining a high true positive (TP) detection ratio. The camera placements selected by our method also provide clear views of the environment for accurate perception.

\subsection{Abrupt Illumination Change}
\hspace*{\parindent} To simulate real-world situations with abrupt illumination changes, such as entering or exiting tunnels, we include a scenario in which the illumination changes abruptly between day and night. 

The performance of different methods in this setting is shown in Fig.~\ref{fig:rapid_change}. Our findings show that methods employing learning-based ACC algorithms predict moderate dynamic parameters, preventing saturation when transitioning from dark to bright scenes. In contrast, the AverageAE method relies solely on the previous frame’s image intensity, often resulting in saturated images during such abrupt changes. Performance for all methods under transitions from bright to dark is comparable to that in constant dark conditions.

\subsection{Additional Qualitative Results on Synthetic Data}
\hspace*{\parindent} We provide further qualitative comparisons between our method and the baselines across all design scenarios using the synthetic images in Fig.~\ref{fig:addition_results}. As discussed in the main paper, our method enables the ACC algorithm to account for non-differentiable effects such as motion blur, which baseline methods fail to model. Specifically, our method predicts lower gain values in high-noise scenarios to improve signal-to-noise ratio (SNR), and it predicts shorter exposure times in conditions with increased motion blur to reduce image degradation. In addition, our method co-designs camera hardware, such as pixel size, to complement these adaptive settings. Larger pixel sizes are chosen when higher noise and motion blur are expected, to increase the number of collected photons for higher SNRs. These design choices allow our method to detect small and distant objects more reliably across both calibrated and challenging scenarios.

\subsection{Qualitative Results on Real-World Data}
\hspace*{\parindent} Fig.~\ref{fig:waymo_results} presents qualitative comparisons between our method and the baselines across varying scenarios in the Waymo Open dataset~\cite{sun2020waymo}. The results show consistent trends with the synthetic experiments and validate the practicality of our method. Our method effectively balances motion blur and SNR, achieving improved task performance across all scenarios, particularly under challenging conditions, owing to the proposed DF-Grad method. Moreover, by jointly optimising camera hardware, our method attains higher SNRs while maintaining sufficient resolution for object detection by changing the pixel size, as reported in the main paper.

